\documentclass[letterpaper]{article} 
\usepackage{aaai25}  
\usepackage{times}  
\usepackage{helvet}  
\usepackage{courier}  
\usepackage[hyphens]{url}  
\usepackage{graphicx} 
\usepackage{natbib}  
\usepackage{caption} 
\usepackage{algorithm}
\usepackage{booktabs}
\usepackage{amssymb}
\usepackage{algorithmic}
\usepackage{amsmath}
\usepackage{bbding}
\usepackage{amsthm,amsmath,amssymb}
\usepackage{mathrsfs}
\usepackage[table]{xcolor}
\usepackage{newfloat}
\usepackage{listings}
\DeclareCaptionStyle{ruled}{labelfont=normalfont,labelsep=colon,strut=off} 
\floatstyle{ruled}
\newfloat{listing}{tb}{lst}{}
\floatname{listing}{Listing}
\title{Mamba YOLO: A Simple Baseline for Object Detection with State Space Model}

\author {
    Zeyu Wang\textsuperscript{\rm 1,\rm 2,\thanks{Equal contribution.}},
    Chen Li\textsuperscript{\rm 1,\rm 2,$^\ast$},
    Huiying Xu\textsuperscript{\rm 1,\rm 2,\thanks{Corresponding author.}}, 
    Xinzhong Zhu\textsuperscript{\rm 1,\rm 2,\rm 3,$^\dagger$},
    Hongbo Li\textsuperscript{\rm 3}
}
\affiliations {
    \textsuperscript{\rm 1}College of Computer Science and Technology, Zhejiang Normal University, Zhejiang, 311231, China\\
    \textsuperscript{\rm 2}Research Institute of Hangzhou Artificial Intelligence, Zhejiang Normal University, Hangzhou, Zhejiang, 311231, China\\
    \textsuperscript{\rm 3}Beijing Geekplus Technology Co, Ltd, Beijing, 100101, China\\
    \{14797857499,LilSodaChen,xhy,zxz\}@zjnu.edu.cn, jason.li@geekplus.com
}

\begin{document}
\maketitle
\begin{abstract}
Driven by the rapid development of deep learning technology, the YOLO series has set a new benchmark for real-time object detectors. Additionally, transformer-based structures have emerged as the most powerful solution in the field, greatly extending the model's receptive field and achieving significant performance improvements. However, this improvement comes at a cost as the quadratic complexity of the self-attentive mechanism increases the computational burden of the model. To address this problem, we introduce a simple yet effective baseline approach called Mamba YOLO. Our contributions are as follows: 1) We propose that the ODMamba backbone introduce a \textbf{S}tate \textbf{S}pace \textbf{M}odel (\textbf{SSM}) with linear complexity to address the quadratic complexity of self-attention.  Unlike the other Transformer-base and SSM-base method, ODMamba is simple to train without pretraining. 2) For real-time requirement, we designed the macro structure of ODMamba, determined the optimal stage ratio and scaling size. 3) We design the RG Block that employs a multi-branch structure to model the channel dimensions, which addresses the possible limitations of SSM in sequence modeling, such as insufficient receptive fields and weak image localization. This design captures localized image dependencies more accurately and significantly. Extensive experiments on the publicly available COCO benchmark dataset show that Mamba YOLO achieves state-of-the-art performance compared to previous methods. Specifically, a tiny version of Mamba YOLO achieves a \textbf{7.5}\% improvement in mAP on a single 4090 GPU with an inference time of \textbf{1.5} ms. The pytorch code is available at: \url{https://github.com/HZAI-ZJNU/Mamba-YOLO}
\end{abstract}

\begin{figure}[h]
\includegraphics[width=0.47\textwidth]{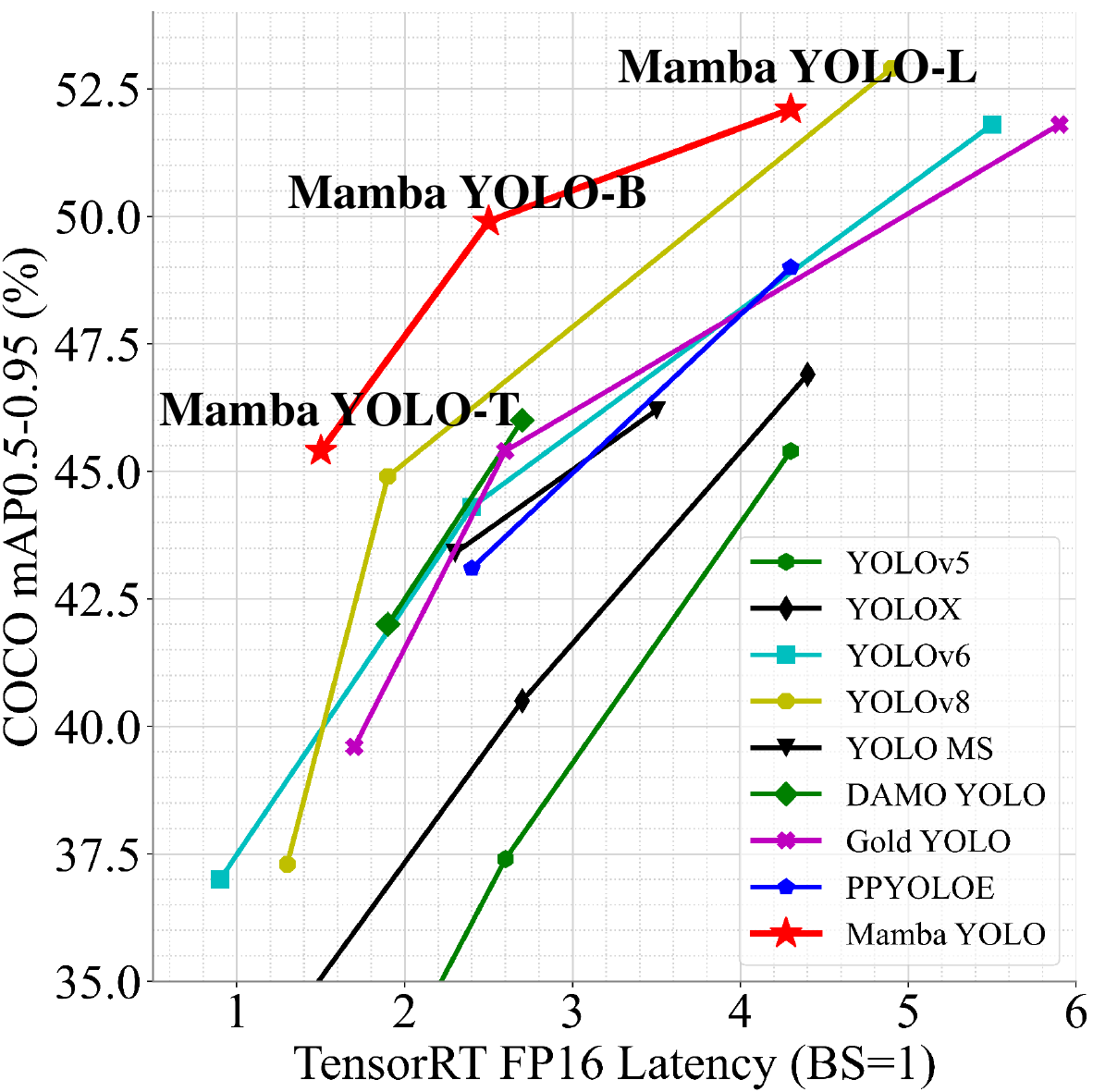}
\caption{Comparisons of the real-time object detecors on the MSCOCO dataset. The object detection method based on SSM achieves the best trade-off between performance and computations.}
\label{fig:sota} 
\end{figure}

\section{Introduction}
In recent years, deep learning has rapidly advanced, particularly in the field of computer vision, where a series of powerful architectures have achieved impressive performance. From \textbf{C}onvolutional \textbf{N}eural \textbf{N}etworks (\textbf{CNNs}) \cite{densely,efficientnet,convnet} to \textbf{Vi}sion \textbf{T}ransformers (\textbf{ViTs}) \cite{swinvit,transnext}, the application of various structures has demonstrated their strong potential in computer vision. In the downstream task of object detection, CNNs \cite{fasterrcnn, SSD} and Transformer structures \cite{detr, dino} are predominantly used. While CNNs and their series of improvements offer fast execution speeds while ensuring accuracy, they suffer from poor image correlation. To address this issue, researchers have introduced ViTs into the field of object detection, such as the DETR series \cite{detr, deformabledetr}, which leverages the powerful global modeling capabilities of self-attention. With hardware advancements, the increased memory computation brought about by this structure does not pose too much of a problem. However, in recent years, more work \cite{convnet, emo, repvit} has begun to rethink how to design CNNs to make models faster, and more practitioners are becoming dissatisfied with the quadratic complexity of Transformer structures. They are starting to use hybrid structures to reconstruct models and reduce complexity, such as MobileVit \cite{mobilevit}, EdgeVit \cite{edgevit}, and EfficientFormer \cite{EfficientFormerV2}. Hybrid models also bring challenges, and the apparent decline in performance is a concern. Therefore, finding a balance between performance and speed has always been a concern for researchers. Recently, methods based on Structured Self-Modulation (SSM), such as Mamba \cite{mamba}, have provided new ideas for solving these problems due to their strong modeling capabilities for long-distance dependencies and the superior properties of linear time complexity.
 \begin{figure*}[h!]
  \centering
  \includegraphics[width=1\textwidth]{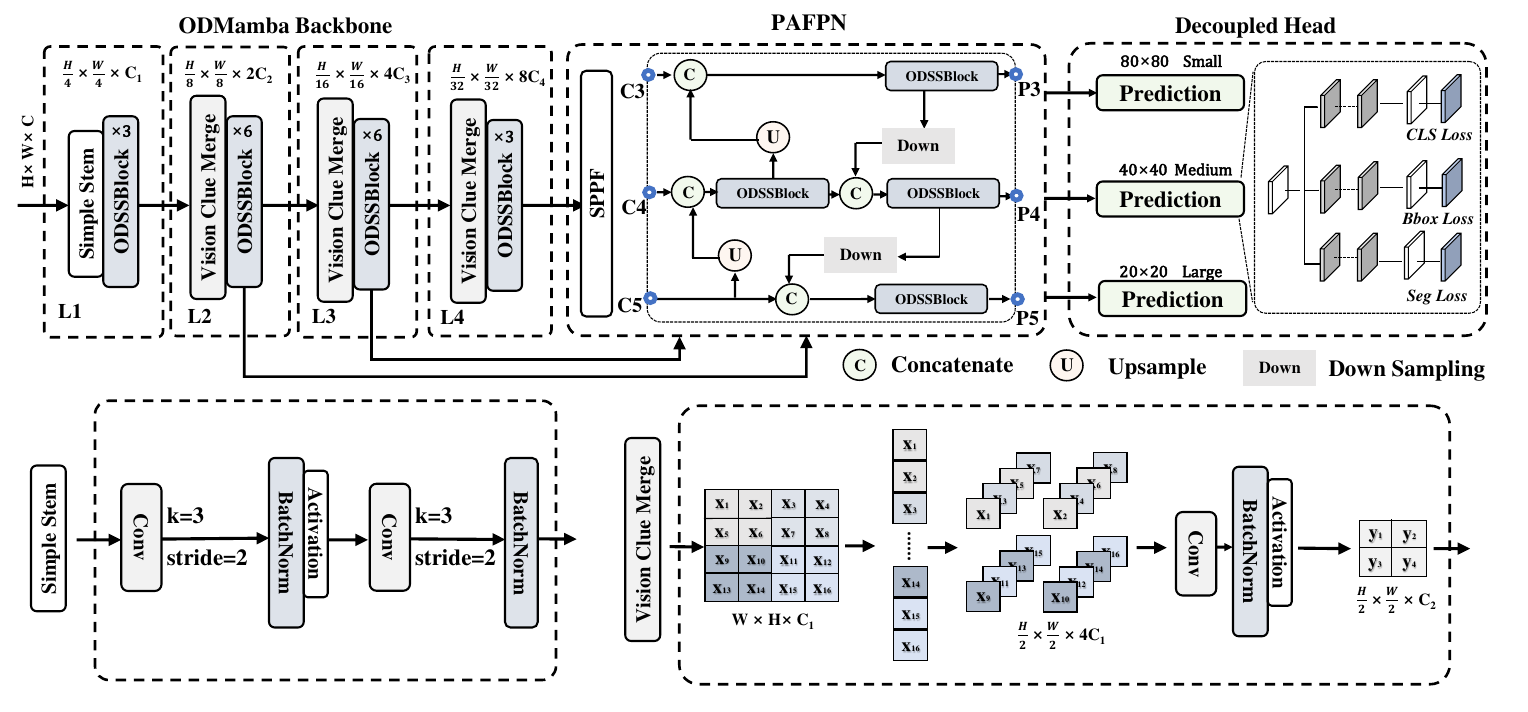}
  \captionsetup{skip=10pt} 
  \caption{Illustration of the Mamba YOLO architecture. Mamba YOLO utilizes the ODSSBlock with selective SSM to construct its backbone, employing a Simple Stem to segment the input image into multiple patches, and using Vision Clue Merge for the downsampling operation.  Multi-level features such as $\{C3,C4,C5\}$ are extracted from the backbone network and then fused into the PAFPN, and high-level semantic features and low-level spatial features are refined and fused by the ODSSBlock, and the resulting $\{P3,P4,P5\}$ features are outputted to the Decoupled Head to output the detection results.}
  \label{fig:Mamba YOLO}
  
\end{figure*}

This paper proposes a detector model called Mamba YOLO. We have designed the Object Detection Structured ODSSBlock module, as shown in Figure \ref{fig:Mamba YOLO}, applying the SSM to the field of object detection. In contrast to the Visual State Space Block \cite{vmamba} used for image classification, object detection tasks often involve images with higher resolution and pixel density. Given that the SSM is primarily designed for textual sequence modeling, it lacks the inherent capability to fully exploit the channel depth present in images. To leverage the enhanced detail and multi-channel information provided by these high-resolution images, we introduced the \textbf{R}esidual \textbf{G}ated (\textbf{RG}) Block architecture. This structure is employed \textbf{S}elective-\textbf{S}can-\textbf{2D} (\textbf{SS2D}) processing to refine the output, utilizing high-dimensional dot product operations to enhance inter-channel correlations and extract richer feature representations. We have conducted exhaustive experiments on MSCOCO \cite{coco}, and the results show that Mamba YOLO is very competitive in general object detection tasks on MSCOCO. The main contributions of this paper can be summarized as follows:
\begin{itemize}
\item Our propose SSM-based Mamba YOLO has a simple and efficient structure with linear memory complexity and does not require pre-training on large-scale datasets, establishing a new baseline for YOLO in object detection.
\item We propose the ODSSBlock to compensate for the local modeling capability of SSM. By rethinking the design of the MLP layer, we introduce the RG Block by combining the idea of gated aggregation with effective convolution and residual connectivity, which effectively captures local dependencies and enhances model robustness.
\item We design a set of models, Mamba YOLO (Tiny/ Base/ Large), with different scales to support the deployment of tasks with different sizes and scales. Experiments on MSCOCO, as shown in Figure \ref{fig:sota}, demonstrate that our Mamba YOLO achieves a significant performance improvement compared to existing state-of-the-art approaches.
\end{itemize}

\section{Related Work}
\subsubsection{Real-Time Object Detectors}
Early performance improvements in YOLO were closely related to improvements in the backbone and led to the widespread adoption of DarkNet. YOLOv7 \cite{yolov7} proposes the E-ELAN structure to enhance the model capability without destroying the original.  YOLO8 \cite{yolov8} combines the features of the previous generations of YOLOs and adopts the \textbf{C}SPDarknet53 to \textbf{2}-Stage \textbf{F}PN (\textbf{C2f}) \cite{yolov8} structure with richer gradient streams, which is lightweight and adaptable to different scenarios while taking accuracy into account.  Recently, Gold YOLO \cite{goldyolo} introduced a new mechanism named \textbf{G}ather-and-\textbf{D}istribute (\textbf{GD}), which is realized by self-attention operation to solve the problem of information fusion of traditional feature pyramid networks \cite{fpn} and Rep-PAN \cite{yolov6}, and succeeded in achieve the SOTA. 
\subsubsection{End-to-End Object Detectors}
DETR \cite{detr} introduces Transformer to object detection for the first time, using a transformer encoder-decoder architecture that bypasses traditional handcrafted components like anchor generation and non-maximum suppression, treating detection as a straightforward ensemble predictionproblem.  Deformable DETR \cite{deformabledetr} introduces Deformable Attention, a variant of Transformer Attention for sampling a sparse set of keypoints around a reference location, addressing the limitations of DETR in handling high-resolution feature maps.  DINO \cite{dino}  integrates a hybrid query selection strategy, deformable attention and demonstrated training with injected noise and performance improvement through query optimization.  RT-DETR \cite{rtdetr}  proposed a hybrid encoder to decouple intra-scale interactions and cross-scale fusion for efficient multi-scale feature processing. However, the excellent performance of DETRs relies heavily on pre-training operations on large-scale datasets, and with the challenges of training convergence, computational cost, and small-object detection for DETRs, YOLOs are still the SOTA in the small modeling domain with both accuracy and speed.
\subsubsection{Vision State Space Models}
 Based on the study of SSM \cite{gu2022efficiently,gu2021combining,smith2023simplified}, Mamba \cite{mamba} shows linear complexity in input size, and solves the computational efficiency problem of Transformer on long sequences of modeling state space.  In the field of generalized visual backbone, Vision Mamba \cite{vim}  proposed a pure visual backbone model based on selective SSM, marking the first time that Mamba has been introduced into the field of vision. VMamba \cite{vmamba}  introduced the Cross-Scan module to enable the model to 2D image Selective scanning enhances visual processing and demonstrates superiority on image classification tasks.  LocalMamba \cite{localmamba} focuses on window scanning strategies for visuospatial models, optimizes visual information to capture local dependencies, and introduces dynamic scanning methods to search for optimal choices for different layers. Inspired by the remarkable results achieved by VMamba in the field of visual tasks, this paper presents for the first time Mamba YOLO, a new SSM model that, unlike traditional SSM-based visual backbones, does not require pre-training on large-scale datasets (e.g. ImageNet \cite{ImageNet}, Object365 \cite{Objects365} ), it aims to take into account the global sensory field while demonstrating its potential in object detection.

\section{Method}
\subsection{Preliminaries}

The structured state-space sequence models S4 \cite{gu2022efficiently} and Mamba \cite{mamba}, rooted in SSM, both stem from a continuous system that maps a univariate sequence\begin{math}\ x(t)\in\mathbb{R}\end{math} into an output sequence\begin{math}\ y(t)\end{math} via an implicit latent intermediate state\begin{math}\ h(t)\in\mathbb{R}^N\end{math}
.  This design not only bridges the relationship between inputs and outputs but also encapsulates temporal dynamics.  The system can be mathematically defined as follows:
\begin{equation}
\label{eq:1}
    h'(t) =\mathbf{A}h(t)+\mathbf{B}x(t)
\end{equation}
\begin{equation}
    y(t) =\mathbf{C}h(t)
\end{equation}
In Equation \eqref{eq:1}, $\mathbf{A}\in\mathbb{R}^{N\times N}$ represents the state transition matrix, which governs how the hidden state evolves over time, while $\mathbf{B}\in\mathbb{R}^{N\times1}$ denotes the weight matrix for the input space in relation to the hidden state.  Moreover, $\mathbf{C}\in\mathbb{R}^{N\times1}$ is the observation matrix, which maps the hidden intermediate state to the output.  Mamba applies this continuous system to discrete-time sequence data by employing fixed discretization rules to transform the parameters $\mathbf{A}$ and $\mathbf{B}$ into their discrete counterparts $\overline{\mathbf{A}}$ and $\overline{\mathbf{B}}$, respectively, thereby better integrating the system into deep learning architectures.  A commonly used discretization method for this purpose is the Zero-Order Hold (ZOH).  The discretized version can be defined as follows: 
\begin{equation}
    \overline{\mathbf{A}} = \texttt{exp}(\mathbf{\Delta A})
\end{equation}
\begin{equation}
\label{eq:4}
    \overline{\mathbf{B}} = {(\mathbf{\Delta A})}^{-1}(\texttt{exp}(\mathbf{\Delta A})-\mathbf{I})\mathbf{\Delta B}
\end{equation}
In Equation \eqref{eq:4}, $\mathbf{\Delta}$ represents a time scale parameter that adjusts the temporal resolution of the model, $\mathbf{\Delta A}$ and $\mathbf{\Delta B}$ correspondingly denote the discrete-time counterparts of the continuous parameters over the given time interval.  Here, $\mathbf{I}$ represents the identity matrix.  After transformation, the model computes via linear recursive forms, which can be defined as follows:
\begin{equation}
    h'(t) = \overline{\mathbf{A}}h_{t-1}+\overline{\mathbf{B}}x_t
\end{equation}
\begin{equation}
    y_t = \mathbf{C}h_t
\end{equation}
The entire sequence transformation can also be represented in a convolutional form, which is defined as follows:
\begin{equation}
    \overline{\mathbf{K}} =(\mathbf{C}\overline{\mathbf{B}},\mathbf{C}\overline{\mathbf{AB}},. . . ,\mathbf{C}{\overline{\mathbf{A}}}^{L-1}\overline{\mathbf{B}})
\end{equation}
\begin{equation}
     y = x\ast\overline{\mathbf{K}}
\end{equation}
wherein, $\overline{\mathbf{K}} \in R^{L}$ represents the structured convolutional kernel, with $\mathbf{L}$ denoting the length of the input equence.  In the design presented in this paper, the model employs a convolutional form for parallel training and utilizes a linear recursive formulation for efficient autoregressive inference. 
\subsection{Overall Architecture}
An overview of the architecture of Mamba YOLO is illustrated in Figure \ref{fig:Mamba YOLO}.  Our object detection model is divided into the ODMamba backbone and neck parts.  ODMamba consists of the Simple Stem, Downsample Block.  In the neck, we follow the design of PAFPN \cite{yolov8} using the ODSSBlock module instead of C2f to capture a more gradient-rich information flow.  The backbone first undergoes downsampling through a Stem module, resulting in a 2D feature map with a resolution of $\frac{H}{4}$,$\frac{W}{4}$.  Subsequently, all models consist of ODSSBlock followed by a VisionClue Merge module for further downsampling.  In the neck part, we adopt the design of PAFPN, using ODSSBlock to replace C2f, where Conv is solely responsible for downsampling. 
\paragraph{Simple Stem} 
Modern ViTs typically employ segmented patches as their initial modules, dividing the image into non-overlapping segments.  This segmentation process is achieved through a convolutional operation with a kernel size of 4 and a stride of 4.  However, recent research, such as that from EfficientFormerV2 \cite{EfficientFormerV2}, suggests that this approach may limit the optimization capabilities of ViTs, impacting overall performance.  To strike a balance between performance and efficiency, we propose a streamlined stem layer.  Instead of using non-overlapping patches, we employ two convolutions with a stride of 2 and a kernel size of 3.  
\paragraph{Vision Clue Merge} 
While CNNs and ViTs structures commonly employ convolutions for downsampling, we discovered that this approach interferes with the selective operation of SS2D \cite{vmamba} across different information flow stages.  To address this, VMamba  splits the 2D feature map and reduces dimensions using $1\times1$ convolutions.  Our findings indicate that preserving more visual clues for SSM benefits model training.  In contrast to the conventional halving of dimensions, we streamline this process by: 
\begin{enumerate}
\item Removing the norm. 
\item Splitting the dimension map. 
\item Appending excess feature maps to the channel dimension.  
\item Utilizing a $4\times$ compressed pointwise convolution for downsampling.
\end{enumerate}
Unlike the use of a $3\times3$ convolution with a stride of 2, our method preserves the feature map selected by SS2D from the previous layer.  
\begin{figure}[h!]
  \centering
  \includegraphics[width=0.5\textwidth]{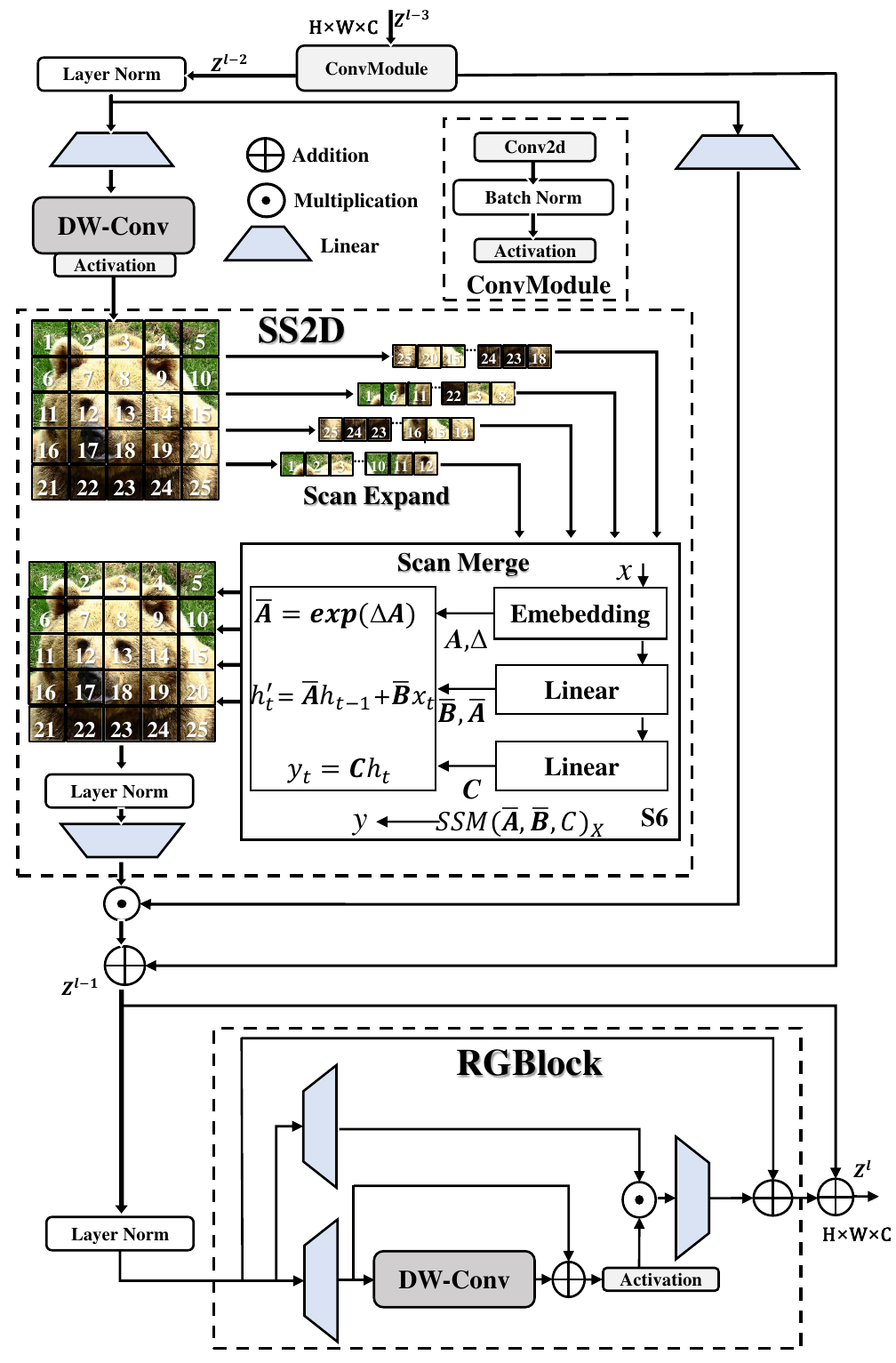}
  \captionsetup{skip=10pt} 
  \caption{Illustration of the ODSSBlock architecture. }
  \label{fig:ODSS Block}
  
\end{figure}

\begin{table*}[h!]
\small
\centering
  \label{tab:commands}
  \begin{tabular}{cccccccccc}
    \toprule
    Method & Params & FLOPs & Latency & ${\rm AP}^{val}(\%)$ & ${\rm AP}_{\mathbf{50}}^{val}(\%)$ & ${\rm AP}_{\mathbf{75}}^{val}(\%)$ & ${\rm AP}_{\mathbf{S}}^{val}(\%)$ & ${\rm AP}_{\mathbf{M}}^{val}(\%)$ & ${\rm AP}_{\mathbf{L}}^{val}(\%)$ \\
    \midrule
    YOLOv6-3.0-N\dag    & 4.7M & 11.7G & 0.9ms & 37.6 & 53.1 & 40.7 & 17.8 & 42.0 & 54.8 \\
    YOLOv8-N        & 3.2M & 8.7G  & 1.3ms & 37.3 & 52.6 & 40.6 & 18.8 & 41.0 & 53.5 \\
    YOLO-MS-XS      & 4.5M & 17.4G & 2.3ms & \underline{43.4} & 60.4 & \underline{47.6} & \underline{23.7} & \underline{48.3} & \underline{60.3} \\
    DAMO YOLO-T\dag     & 8.5M & 18.1G & 1.9ms & 42.0 & 58.0 & 45.2 & 23.0 & 46.1 & 58.5 \\
    PPYOLOE-S*       & 7.9M & 14.4G & 2.4ms & 43.0 & \underline{60.5} & 46.6 & 23.2 & 46.4 & 56.9 \\
    Gold-YOLO-N\dag     & 5.6M & 12.1G & 1.7ms & 39.9 & 55.9 & 44.5 & 19.7 & 44.1 & 57.0 \\
    \rowcolor{gray!20}
    Mamba YOLO-T    & 5.8M & 13.2G & 1.5ms & \textbf{44.5} & \textbf{61.2} & \textbf{48.2} & \textbf{24.7} & \textbf{48.8} & \textbf{62.0} \\
    \midrule
    YOLOv6-3.0-S\dag    & 18.5M & 45.3G & 2.4ms & 45.1 & 62.1 & 48.7 & 24.8 & 50.3 & 62.6 \\
    YOLOv8-S        & 11.2M & 28.6G & 1.9ms & 44.9 & 61.8 & 48.6 & 26.0 & 49.9 & 61.0 \\
    YOLO-MS-S       & 8.1M  & 31.2G & 3.5ms & 46.2 & \underline{63.7} & 50.5 & 26.9 & 50.5 & 63.0 \\
    DAMO YOLO-S\dag     & 12.3M & 37.8G & 2.7ms & 46.0 & 61.9 & 49.5 & 25.9 & 50.6 & 62.5 \\
    PPYOLOE-M*      & 23.4M & 49.9G & 4.3ms & \underline{49.0} & \textbf{66.5} &  \underline{53.0} &  \underline{28.6} & \underline{52.9} & \underline{63.8} \\
    Gold-YOLO-S*     & 21.5M & 46.0G & 2.9ms & 45.4 & 62.5 & 49.2 & 25.3 & 50.2 & 62.6 \\
    \rowcolor{gray!20}
    Mamba YOLO-B    & 19.1M & 45.4G & 2.2ms  & \textbf{49.1} & \textbf{66.5} & \textbf{53.5}& \textbf{30.6} & \textbf{54.0} & \textbf{66.4} \\
    \midrule
    YOLOv6-3.0-L\dag    & 59.6M & 150.7G & 5.5ms & 51.8 & 69.2 & \textbf{57.7} & \underline{34.7} & \underline{58.2} & \underline{69.5} \\
    YOLOv8-L        & 43.7M & 165.2G & 4.9ms & \textbf{52.9} & \textbf{69.8} & \textbf{57.7} & \textbf{35.5} & \textbf{58.5} & \textbf{69.8} \\
    Gold-YOLO-L*     & 75.1M & 151.7G & 5.9ms & 51.8 & 68.9 & \underline{57.6} & 34.1 & 57.4 & 68.2 \\
    DINO-R50*        & 47.0M & 279.0G &   51.2ms    & 50.9 & 69.0 & 55.3 & 34.6 & 54.1 & 64.6 \\
    \rowcolor{gray!20}
    Mamba YOLO-L    & 57.6M & 156.2G & 4.3ms & \underline{52.1} & \textbf{69.8} & 56.5 & 34.1 & 57.3 & 68.1 \\
    \bottomrule
  \end{tabular}
\caption{Comparison of Mamba YOLO with other detectors on the MSCOCO val. To ensure a fair comparison, all models use official pre-trained models tested for latency on NVIDIA 4090 GPU using the half-precision floating-point format (FP16), with the TensorRT version 8.4.3 and cuDNN version 8.2.0. `\dag'denotes that additional self-distillation is performed after the completion of training. `*' indicates the use of ImageNet or a similarly large-scale object detection dataset for supervised pre-training. The results pertaining to the proposed Mamba YOLO model are highlighted in gray. The best and second-best results are bold faced and underlined, respectively.}
\label{SOTA}
\end{table*}
\captionsetup[figure]{skip=10pt}

\subsection{ODSSBlock}
As shown in Fig. \ref{fig:ODSS Block}, the ODSSBlock is the core module of Mamba YOLO, and in the input phase it passes through a ConvModule that enables the network to learn a deeper and richer representation of the features, we assume that the shape of the input feature $Z^{l-3}$ is in the shape $\mathbb{R}^{C\times H\times W}$, we have:
\begin{equation}    
    Z^{l-2} = \sigma \left( BatchNorm\left( ConvModule (Z^{l-3}) \right) \right)
\end{equation}
where ${\mathrm{\sigma}}$ denotes the activation function (nonlinear SiLU).  The Layer Normalization and residual linking design of the ODSSBlock draws on a transformer Blocks style architecture, which allows the model to flow efficiently and training in the presence of deep stacking.
\begin{equation}
    Z^{l-1} = SS2D\left(LayerNorm(Z^{l-2}) \right) + Z^{l-2}
\end{equation}
\begin{equation}
   Z^{l} = RG Block \left( LayerNorm(Z^{l-1}) \right) + Z^{l-1}
\end{equation}
ODSSBlock can be decoupled into two separate functional components $SS2D(\cdot)$ and $RG Block(\cdot)$ for global spatial information dissemination and channel information dissemination, where $Z^{l-1}$ denotes the intermediate state following the SS2D).   
\subsubsection{SS2D} Scan Expansion, S6 Block and Scan Merge are the three main steps of the SS2D algorithm, and its main flow is shown in Figure \ref{fig:ODSS Block}.  The scan expansion operation expands the input image into a series of subimages, each of which denotes a specific direction, and when observed from the diagonal viewpoint, the scan expansion operation proceeds along the four symmetric directions, which are top-down, bottom-up, left-right, and word-right-to-left, respectively.  Such a layout not only comprehensively covers all regions of the input image, but also enhances the efficiency and comprehensiveness of multi-dimensional capturing of image features by providing a rich multi-dimensional information base for subsequent feature extraction through systematic direction transformation.  The scan merge operation in SS2D takes the obtained sequences as inputs to the S6 block \cite{mamba} and merges the sequences from the different directions so that the features are extracted to the global features. 

\subsubsection{RG Block}
The original MLP is still the most widely adopted, and the MLP in VMamba architecture also follows the Transformer design, which performs nonlinear transformations on the input sequences to enhance the expressive power of the model.  Recent studies, Gated MLP \cite{gated,gatedmlp} shows strong performance in natural language processing, and we find that the mechanism of gating has the same potential for vision.  In Figure \ref{fig:ODSS Block}, this paper proposes that the simple design of the Residual Gated Block aims to improve the performance of the model at a lower computational cost, and that the RG Block creates two branches from the inputs $f_{A}'$ and $f_{B}'$ to retain global and local information, respectively, $\mathcal{T}(\cdot)$ denotes the linear layer.

\begin{equation}
    \mathcal{R}_{\text{local}}^{l-1} = \mathcal{T}_{\text{local}}^{l-1}(f_{A}') 
\end{equation}
\begin{equation}
    \mathcal{R}_{\text{global}}^{l-1} = \mathcal{T}_{\text{global}}^{l-1}(f_{B}')
\end{equation}
Depth separable convolution is used as the Position Encoding Module on the branch of $\mathcal{R}_{\text{global}}^{l-1}$, and the gradient is more efficiently refluxed during training by means of residual concatenation, which has a lower computational cost and significantly improves the performance by preserving and utilizing the spatial structural information of the image.  The RG Block adopts a nonlinear GeLU as an activation function to control the flow of information at each level. The $\mathcal{Y}(x)$ process can be written as:
\begin{equation}
\mathcal{Y}(x) = \Phi(DWConv(x) \oplus x)
\end{equation}
The local information passing through $\mathcal{Y}(x)$ is multiplied with the global information of $\mathcal{R}_{global}^{l-1}$, the global features are refined by a linear layer to fuse the information of the local channels, and the residual connection is allowed to be summed with the original input of $f_{A}'$ and the features of the hidden layer. The RG Block captures more global and local features while incurring only a slight increase in computational cost, and the resulting output feature $f_{RG}$ is defined as follows:

\begin{equation}
\mathcal{R}_{\text{fusion}}^{l} = \mathcal{R}_{\text{global}}^{l-1} \odot \mathcal{Y}(\mathcal{R}_{\text{local}}^{l-1})
\end{equation}
\begin{equation}
f_{RG} = \mathcal{T}_{\text{fusion}}^{l}(\mathcal{R}_{\text{fusion}}^{l})\oplus f_{A}'
\end{equation}
where $\mathrm{\Phi}$ denotes the activation function (nonlinear GELU).  In this paper.  the gating mechanism in RG Block preserves the spatial information by integrating the convolution operation while making the model more sensitive to the fine-grained features in the image.  Compared with the traditional MLP, RG Block transfers the global dependencies and global features to each pixel to capture the dependencies of the neighboring features, which makes the contextual information rich to further enhance the model's expressive ability. 

\section{Experiments}
In this section, we conduct comprehensive experiments on Mamba YOLO for object detection task.  We employ the MSCOCO dataset to validate the superiority of the proposed Mamba YOLO Comparison with state-of-the-arts. All our models are trained on 8 NVIDIA H800 GPUs.
\begin{figure*}[htbp]
  \centering
  \includegraphics[width=1\textwidth]{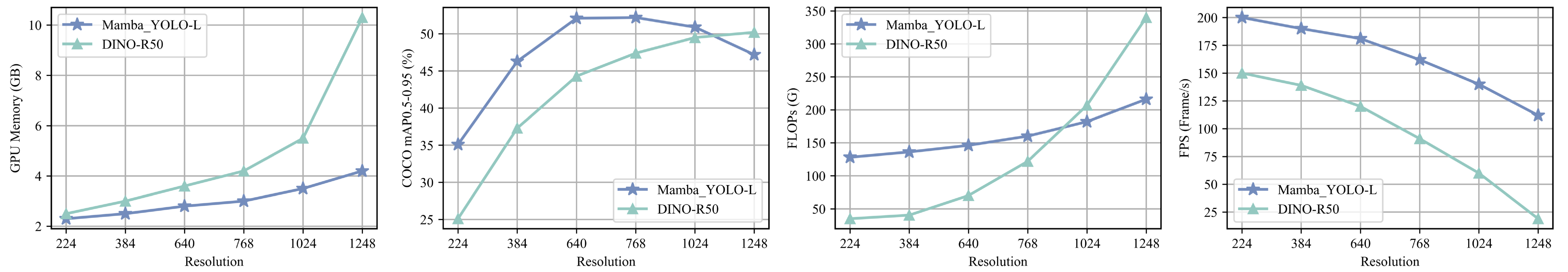}
  \captionsetup{skip=10pt} 
  \caption{Comparison between DINO-R50 and Mamba YOLO-L in terms of GPU memory efficiency and mAP. As the input image resolution increases, DINO requires higher resolution to maintain a high mAP and shows a quadratic growth trend in both GPU memory and FLOPs. In contrast, MambaYOLO maintains a linear increase in GPU memory requirements and achieves the highest performance at a smaller resolution of 640×640, with fewer FLOPs and faster inference.}
  \label{fig:Resolution}
  
\end{figure*}
\begin{figure*}[htbp]
  \centering
  \includegraphics[width=1\textwidth]{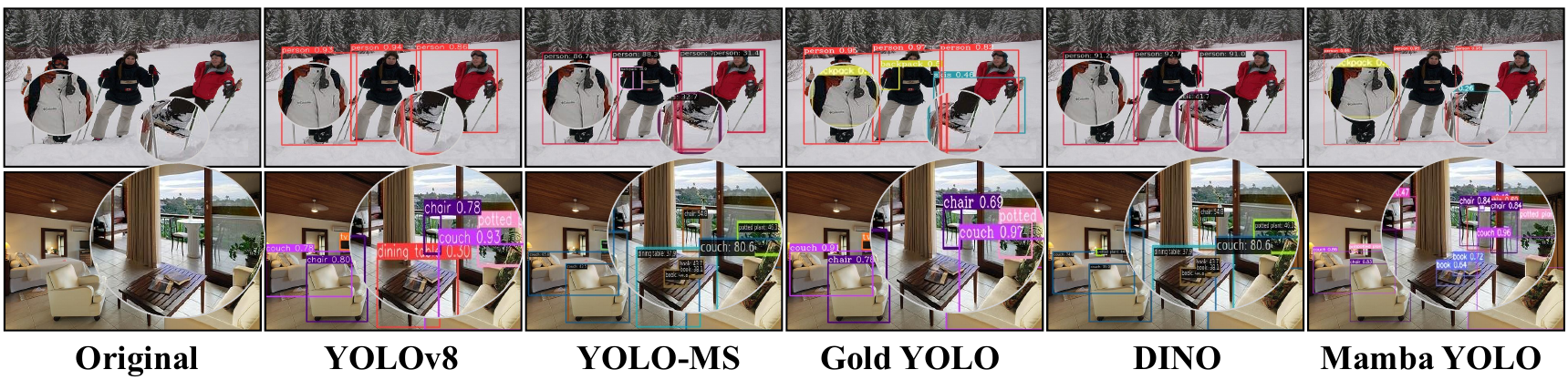}
    \captionsetup{skip=10pt} 
  \caption{Inference results for each detector on the COCO dataset. Detailed objects have been enlarged for better illustration.}
  \label{fig:detect}
\end{figure*}
\subsubsection{Comparison with state-of-the-arts}
Table~\ref{SOTA} illustrates the results of MSCOCO val, demonstrating that our proposed method achieves the best overall trade-off between FLOPs, number of parameters, and accuracy, alongside measured GPU latency. Specifically, compared to high-performing miniature lightweight models such as PPYOLOE-S \cite{ppyolo} /YOLO-MS-XS \cite{yoloms}, Mamba YOLO-T shows a significant increase in AP of 1.1\%/1.5\%, with a reduction in GPU inference latency of 0.9ms/0.2ms. When compared to the baseline model YOLOv8-S with similar accuracy, Mamba YOLO-T reduces the number of parameters by 48\% and FLOPs by 53\%, while decreasing GPU inference latency by 0.4ms. 

Mamba YOLO-B, when compared to Gold-YOLO-M which has a similar number of parameters and FLOPs, achieves an AP gain of 3.7\% higher. Even when compared to PPYOLOE-M with comparable accuracy, Mamba YOLO-B reduces the number of parameters by 18\% and FLOPs by 9\%, with a decrease in GPU inference latency of 1.8ms. For larger models, Mamba YOLO-L also achieves better or comparable performance across all advanced object detectors. Compared to the top-performing Gold-YOLO-L \cite{goldyolo}, Mamba YOLO-L increases AP by 0.3\% with a reduction in the number of parameters by 0.9\%. As you can see from this table, Mamba YOLO-T, which uses the scratch training method, performs better than all the other training methods.

Furthermore, Figure~\ref{fig:Resolution} compares Mamba YOLO-L with DINO-R50 in terms of \textbf{f}rames \textbf{p}er \textbf{s}econd (\textbf{FPS}) and GPU memory usage, showing that Mamba YOLO-L maintains better precision and speed at increased resolutions, with linear growth in memory efficiency and FLOPs. These comparative results demonstrate that, across different scales of Mamba YOLO, our proposed models have notable advantages over existing state-of-the-art methods.
\begin{table}[htbp]
\centering
\resizebox{1.0\linewidth}{!}{%
\label{Ablation study on Mamba YOLO}
\begin{tabular}{ccccc}
\toprule
\small
ODSSBlock & RG Block&Clue Merge &${\rm AP}^{val}(\%)$&${\rm AP}_{\mathbf{50}}^{val}(\%)$\\
\midrule
 & & &    37. 3  &52. 6   \\
  \checkmark & & &  43. 1 & 59. 2  \\
  & \checkmark & & 37. 9 & 53. 9 \\
 & & \checkmark &  35. 6 &50. 9  \\
 \checkmark & &  \checkmark&  43. 4 &59. 5  \\
  \checkmark &  \checkmark & & 44. 1  &60. 1  \\
 \checkmark &  \checkmark &  \checkmark &   44. 5 & 61. 2   \\

\bottomrule
\end{tabular}}
\caption{Ablation study on Mamba YOLO.}
\label{Ablation study on Mamba YOLO}
\vspace{-1em} 
\end{table}

\subsubsection{Ablation study on Mamba YOLO}
In this section, we examine each module in the ODSSBlock independently, and without Clue Merge, we downsample using the traditional convolutional to assess the impact of Vision Clue Merge on accuracy. Mamba YOLO is performed on the MSCOCO dataset to conduct Ablation experiments , and the test model is Mamba YOLO-T.  Table~\ref{Ablation study on Mamba YOLO} of our results shows that cue merging preserves more visual cues for the SSM and also provides evidence for the assertion that the ODSSBlock structure is indeed optimal.

\begin{figure}[h]
  \centering
  \includegraphics[width=0.48\textwidth]{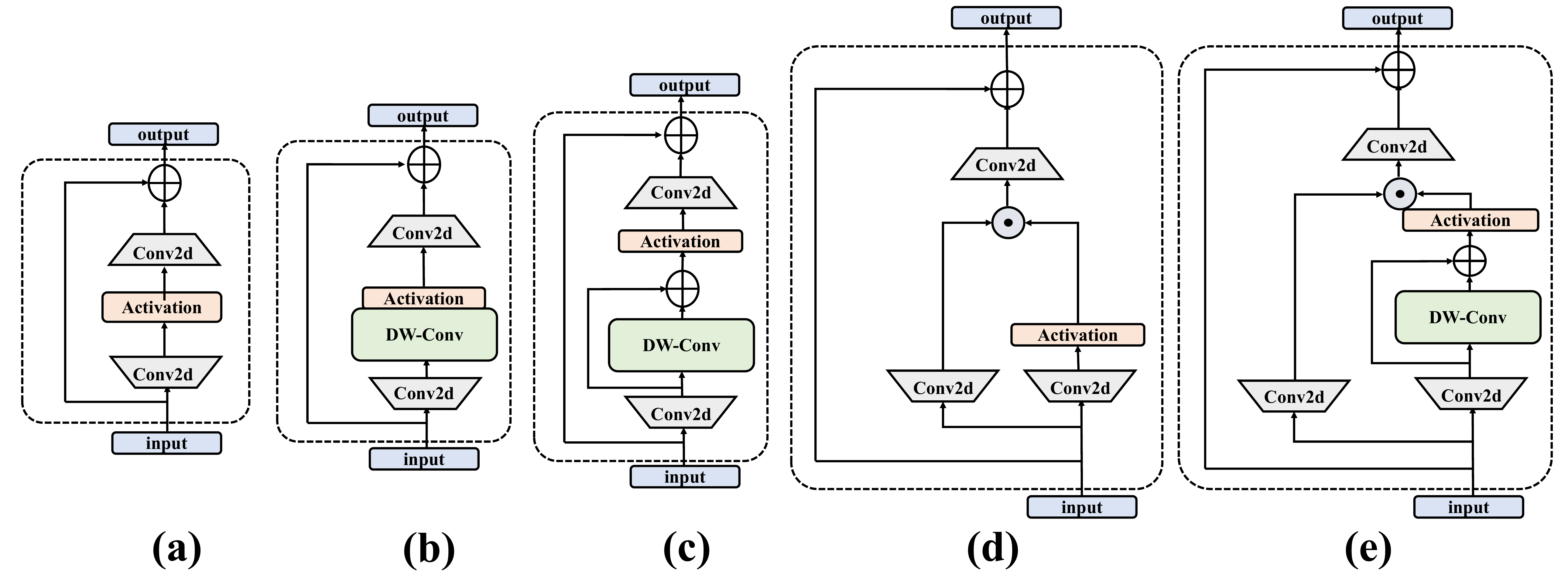}
  \captionsetup{skip=10pt} 
  \caption{RG Block integration designs explored in the ablation study.}
  \label{fig:RG Block}
  \vspace{-1em} 
\end{figure}

\subsubsection{Ablation study on RG Block structure}
RG Block to capture pixel-by-pixel local dependencies by taking global dependencies and global features from pixel-by-pixel. RG Block uses a multi-branch structure to model the channel dimension, which addresses the limitations of SSM in sequence modeling in terms of insufficient sensory fields and weak image localization. Regarding the design details of RG Block, we also consider three variants:
\begin{enumerate}
\item Convolutional MLP, which adds DW-Conv to the original MLP. 
\item Res-Convolutional MLP, which adds DW-Conv in a residual concatenated fashion added to the original MLP. 
\item Gated MLP, an MLP variant designed under the gating mechanism.  
\end{enumerate}
Figure \ref{fig:RG Block} illustrates these variants, and Table~\ref{Ablation study on MPL variants and RG Block. } shows the performance of the original MLP, RG Block, and each variant in the MSCOCO dataset to verify the validity of our analysis on the MLP, with the test model Mamba YOLO-T.  We observe that the introduction of convolution singly does not lead to an effective improvement in the performance, whereas in the variant Figure \ref{fig:RG Block} Gated MLP, its output consists of two linear projections of element multiplication, one of which consists of residual-connected DWConv and gated activation functions, which in fact endows the model with the ability to propagate important features through the hierarchical structure and effectively improves the accuracy and robustness of the model.  This experiment shows that the improvement of the performance of the introduced convolution when dealing with complex image tasks is quite relevant to the gated aggregation mechanism, provided that they are applied in the context of residual connectivity. 
\begin{table}[htbp]
\centering
\small
\resizebox{1.0\linewidth}{!}{%
\begin{tabular}{ccccc}
\toprule
variants&Model & ${\rm AP}^{val}(\%)$&${\rm AP}_{\mathbf{50}}^{val}(\%)$& FLOPs\\
\midrule
(a) &Original          &  43. 0 & 59. 6 & 13.2G\\
(b) &Convolutional       & 43. 2 &  60. 1  &13.3G\\
(c) &Res-Convolutional& 43. 3 &  60. 5 & 13.3G\\
(d) &Gated             &  44. 0 & 60. 8   &13.2G\\
(e) &RG Block(Ours)        &   44. 5 & 61. 2  & 13.2G\\
\bottomrule
\end{tabular}}
\caption{Ablation study on MLP variants and RG Block. The test model is Mamba YOLO-T.}
\label{Ablation study on MPL variants and RG Block. } 
\vspace{-1em} 
\end{table}

\subsubsection{Ablation study on the type of value setting in Mamba YOLO variants}
We explore four different configurations of the number of repetitions of the ODSSBlock in the backbone: $[9,\,3, \,3, \,3]$ imposes an additional computational overhead, but does not result in a corresponding degree of accuracy improvement. $[3,\,9, \,3,\,3]$, $[3,\,3, \,9,\,3]$ and $[3,\,3, \,3,\,9]$, which is actually a redundancy due to excessive duplication of ODSSBlock. Experiments prove that $[3,\,6, \,6,\,3]$ is a more reasonable configuration in Mamba YOLO. In the Neck part, although removing the ODSSBlock can realize a more lightweight model, this will inevitably reduce the accuracy of the model, and the ODSSBlock in the Neck part can provide rich gradient flow and feature fusion. Choosing the output Feature Map to be the $\{P2, P3, P4, P5\}$ variant significantly improves accuracy, but inevitably increases GFLOPs significantly. Mamba YOLO ultimately chose $Blocks$=$[3, 6, 6, 3]$, Feature Map=$\{P3, P4, P5\}$ and used ODSSBlock in the Neck section. This configuration achieves a better balance between accuracy and complexity. better balance between accuracy and complexity, and is more suitable for efficiently performing the instance segmentation task. The results are shown in Table ~\ref{tab:more}.
\begin{table}[htbp]
\centering
\small
\resizebox{1.0\linewidth}{!}{%

  \centering
  \begin{tabular}{cccccc}
    \toprule
   $Blocks$&w/o $SSM_{Neck}$&$Feature Map$ &${\rm AP}^{val}(\%)$&${\rm AP}_{\mathbf{50}}^{val}(\%)$& FLOPs\\
\midrule
$[9, \,3, \,3, \,3]$&\Checkmark & \{P3, P4, P5\} &44. 0&61. 5 &13. 3G\\
$[3, \,9, \,3, \,3]$&\Checkmark& \{P3, P4, P5\} & 43. 4  &60. 9 &13. 2G\\
$[3, \,3, \,9, \,3]$&\Checkmark & \{P3, P4, P5\} &43. 6 &61. 0 & 13. 2G \\
$[3, \,3, \,3, \,9]$&\Checkmark & \{P3, P4, P5\} &43. 8& 61. 0&13. 1G\\
$[3, \,6, \,6, \,3]$&\XSolidBrush &\{P3, P4, P5\}& 42. 2 &60. 3 &11. 4G \\
$[3, \,6, \,6, \,3]$&\Checkmark & \{P2, P3, P4, P5\}& 45. 8&62. 3 &19. 7G\\
$[3, \,6, \,6, \,3]$& \Checkmark&\{P3, P4, P5\} &  44. 5 & 61. 2  & 13. 2G\\
\bottomrule
  \end{tabular}}
  \caption{Ablation study on the type of value setting in Mamba YOLO variants. $Blocks$ indicate the number of repetitions of ODSSBlock in the backbone. \Checkmark indicates that the SSM is used in the neck, \XSolidBrush indicates that the SSM is not used in the neck. In output Feature Map size, P2$=$20×20, P3$=$40×40, P4$=$80×80, P5$=$160×160.}  \label{tab:more}
  \vspace{-1em} 
\end{table}
\subsubsection{Visualization}
To further confirm the advantages of our proposed detection framework, we randomly selected two samples from MSCOCO. Figure~\ref{fig:detect} shows the visualization results of each mainstream detector with Mamba YOLO, and it can be seen that Mamba YOLO is able to achieve accurate detection under a variety of difficult conditions and shows strong ability in detecting highly overlapping, heavily occluded objects with complex backgrounds, and also shows strong ability in detecting highly overlapping and severely occluded objects.
\section{Conclusion}
In this paper, we propose a detector designed based on the SSM and extended by YOLO, the training process is notably simple, as it does not necessitate pre-training on extensive datasets. We re-analyze the limitations of the traditional MLP and propose the RG Block, whose gating mechanism and deep convolutional residual connectivity are designed to give the model the ability to propagate important features in the hierarchical structure.  Our goal is to establish a new baseline of YOLOs, proving that Mamba YOLO are highly competitive.  Our work is the first exploration of the Mamba architecture in the real-time object detection task, and we also hope to bring new ideas to researchers in the field. 
\section{Acknowledgement}
This work was supported by the National Natural Science Foundation of China (62376252); Key Project of Natural Science Foundation of Zhejiang Province (LZ22F030003); Zhejiang Province Leading Geese Plan (2024C02G1123882).
\bibliography{aaai25}

\end{document}